\documentclass[letterpaper, 10 pt, conference]{ieeeconf} 
\IEEEoverridecommandlockouts
\usepackage{cite}
\usepackage{amsmath,amssymb,amsfonts}
\usepackage{algorithmic}
\usepackage{graphicx}
\usepackage{textcomp}
\usepackage{xcolor}
\usepackage{comment}
\usepackage{mathtools}
\usepackage{soul}
\usepackage{url}
\usepackage{dblfloatfix}
\usepackage{amssymb}
\usepackage{subcaption}
\def\BibTeX{{\rm B\kern-.05em{\sc i\kern-.025em b}\kern-.08em
    T\kern-.1667em\lower.7ex\hbox{E}\kern-.125emX}}
\begin{document}

\title{\LARGE \bf
Simulation-based Analysis of a Novel Loop-based Road Topology for Autonomous Vehicles}

\begin{comment}
\author{

\IEEEauthorblockN{
  Stefan Ramdhan\IEEEauthorrefmark{1}
  Winnie Trandinh\IEEEauthorrefmark{1},
  Sathurshan Arulmohan\IEEEauthorrefmark{1},
  Xiayong Hu\IEEEauthorrefmark{2}, \\
  Spencer Deevy\IEEEauthorrefmark{1},
  Victor Bandur\IEEEauthorrefmark{1},
  Vera Pantelic\IEEEauthorrefmark{1},
  Mark Lawford\IEEEauthorrefmark{1},
  Alan Wassyng\IEEEauthorrefmark{1}
}

\IEEEauthorblockA{
	\IEEEauthorrefmark{1}
	\textit{McMaster Centre for Software Certification, McMaster University,  Hamilton, ON, Canada.} \\
	Email: \{ramdhans, trandint, arulmohs, deevys, bandurvp, pantelv, lawford, wassyng\}@mcmaster.ca 
	\\
	\IEEEauthorrefmark{2}
	\textit{HU SUNWAY Inc.,, ON, Canada.} 
	Email: jason.hu68@gmail.com
}
\end{comment}

\author{Stefan Ramdhan$^{1}$, Winnie Trandinh$^{1}$, Sathurshan Arulmohan$^{1}$, Xiayong Hu$^{2}$, Spencer Deevy$^{1}$, Victor Bandur$^{1}$,\\ Vera Pantelic$^{1}$, Mark Lawford$^{1}$, Alan Wassyng$^{1}$% <-this % stops a space
%\thanks{*This work was not supported by any organization}% <-this % stops a space
\thanks{$^{1}$McMaster Centre for Software Certification, Department of Computing and Software,
        McMaster University, Hamilton ON, Canada.
        {\tt\small \{ramdhans, trandint, arulmohs, deevys, bandurvp, pantelv, lawford, wassyng\}@mcmaster.ca}}%
\thanks{$^{2}$Xiayong Hu is with HU SUNWAY Inc., Markham, ON, Canada.
        {\tt\small husunwayinc@gmail.com}}%
}

\maketitle

\thispagestyle{empty}
\pagestyle{empty}

\begin{abstract}
    The challenges in implementing SAE Level 4/5 autonomous vehicles are manifold, with intersection navigation being a pervasive one. 
    We analyze a novel road topology invented by a co-author of this paper, Xiayong Hu. The topology eliminates the need for traditional traffic control and cross-traffic at intersections, potentially %simplifies the design and implementation of autonomous driving systems, and also 
    improving the safety of autonomous driving systems.  
    The topology, herein called the Zonal Road Topology, consists of unidirectional loops of road with traffic flowing either clockwise or counter-clockwise. Adjacent loops are directionally aligned with one another, allowing vehicles to transfer from one loop to another through a simple lane change. 
    To evaluate the Zonal Road Topology, a one km\textsuperscript{2} pilot-track near Changshu, China is currently being set aside for testing. In parallel, traffic simulations are being performed. To this end,
    we conduct a simulation-based comparison between the Zonal Road Topology and a traditional road topology for a generic Electric Vehicle (EV) using the Simulation for Urban MObility (SUMO) platform and MATLAB/Simulink. 
    We analyze the topologies in terms of their travel efficiency, safety, energy usage, and capacity. Drive time, number of halts, progress rate, and other metrics are analyzed across varied traffic levels to investigate the advantages and disadvantages of the Zonal Road Topology. 
    Our results indicate that vehicles on the Zonal Road Topology have a lower, more consistent drive time with greater traffic throughput, while using less energy on average. 
    %The Zonal Road Topology also has a greater, more consistent traffic throughput. 
    These results become more prominent at higher traffic densities.
\end{abstract}


\section{Introduction}
\label{sec:introduction}
%\subsection{Problem Statement \& Motivation.}

%BEGINNING

Fully autonomous vehicles have the potential to improve the safety and efficiency of traffic on our roads, primarily by eliminating driver error. Many predictions have Society of Automotive Engineers Level 5 (L5) autonomy being introduced by 2030 - 2050, with an even longer time horizon to achieve wide-scale adoption within the whole vehicle fleet \cite{martinezdiaz}. 
This wide-scale adoption is, in part, hindered by the complexity of driving. A safe driver must be aware of their surroundings and react swiftly to many stimuli: other vehicles, pedestrians, road signs, traffic lights, etc. Current Advanced Driver Assistance Systems (ADAS) still need improvement; one study shows that at higher speeds, ADAS systems fail to detect and react to pedestrians more frequently as speeds increase \cite{AAA}.

Current road layouts and traffic signals, initially developed for carriage travel and subsequently adapted for modern vehicles, present significant obstacles in achieving L5 autonomous driving. These roads, repeatedly modified over time, were conceived with human drivers in mind.  
Intersections are particularly challenging, requiring complex decision-making from both human drivers and autonomous vehicles.
Approximately 40\% of the 5,811,000 crashes that occurred in the United States in 2008 were intersection-related, with 96\% of those having the critical cause attributed to the driver \cite{choi}. Not only are intersections dangerous, they are also inefficient and costly \cite{loos}, especially as traffic scales \cite{shi} \cite{azimi}.
Roundabouts are a  first step toward reducing fatal and injury accidents at road intersections \cite{elvik}, increasing throughput \cite{demir} \cite{zhou_longfei}, and reducing emissions under low and high traffic conditions when replacing signalized intersections \cite{ARIUS}. 
However, a more radical change is needed to facilitate full L5 autonomy. The use of Dedicated Lanes (DL) for Connected and Autonomous Vehicles (CAVs) is one such approach, exemplified by Japan's upcoming 100 kilometer lane for self-driving trucks \cite{Jen2024}. Still, research suggests that merely allocating specific lanes for autonomous vehicles could be insufficient for a comprehensive solution\cite{he2022}\cite{chen2022}. An effective DL policy needs to factor in elements such as the Market Penetration Rate (MPR) of CAVs to enhance performance and throughput \cite{he2022}, which in turn could influence emission outcomes \cite{chen2022}. 

While the current research predominantly focuses on retrofitting existing road infrastructures to support autonomous vehicles, creating simplified and dedicated road topologies for autonomous driving could enable current technologies to achieve higher or even full automation. This approach presents the opportunity to build entirely new smart cities from scratch, fostering safer travel, reduced traffic congestion, and facilitating the full implementation of autonomous driving systems.

In this paper, we analyze a novel road topology for autonomous vehicles, herein called the Zonal Road Topology, invented by Xiayong Hu, patented under US 2021/0404123A1 \cite{patent}. Drawing inspiration from the ancient Chinese philosophy of Yin and Yang balance, this concept parallels the traditional Chinese medicinal belief that a body free from blockages is pain-free, underlining that unobstructed circulation and balance are vital for health. This principle is applied to urban traffic design, advocating for future cities to be free from congestion. The Zonal Road Topology offers a transformative vision for the smart cities of the future, marked by several key enhancements:
\begin{itemize}
\item \textbf{Seamless Mobility:} Envision a city with a constant flow of traffic, free from the snarls of rush hour, offering a novel point-to-point travel experience with highly predictable timing.

\item \textbf{Enhanced Safety:} Approximately 40\% of traffic accidents occur due to conflicts at intersections \cite{choi}. The implementation of the Zonal Road Topology is projected to significantly lower this statistic.
\item \textbf{Simplified Autonomous Decision-Making:} By transforming classic intersection conflicts into simpler lane-changing scenarios, this method streamlines the decision-making process for autonomous driving systems.
\end{itemize}

Together, these advancements pave the way for shared mobility solutions, in synergy with autonomous vehicles, promoting a more eco-friendly urban environment. Cities could recover vast stretches of land currently dedicated to parking due to more optimized space utilization, transforming them into parks, recreational areas, and other forms of green space \cite{duarte}.

%\subsection{Contributions \& Outline of this Paper}

This paper presents the first experimental analysis of the proposed road topology. It is a preliminary simulation-based comparative analysis of the Zonal Road Topology and traditional road topologies with static and adaptive traffic light scheduling. We perform our analysis using the Simulation for Urban MObility (SUMO) \cite{sumo} traffic simulation software and an Electric Vehicle Model in Simulink, comparing the topologies in terms of their travel efficiency, safety, energy usage, and capacity.
%We analyze metrics defined in Section \ref{sec:metrics}.

The rest of the paper is structured as follows. Section \ref{sec:litReview} contains a review of work related to traffic simulation. Section \ref{sec:topology} explains the Zonal Road Topology in greater detail. Section \ref{sec:methodology} contains a detailed description of the simulation and experiment methodology. Section \ref{sec:metrics} contains a detailed description of the metrics used. Section \ref{sec:results} contains the data collected and an in-depth analysis. Finally, we conclude with Section \ref{sec:conclusion} and discuss future work.

%\section{Introduction}
%This document is a model and instructions for \LaTeX.
%Please observe the conference page limits. 

\section{Related Work}
\label{sec:litReview}

Microscopic traffic simulators are typically used to assess the performance of road topologies, especially to analyze how roundabouts perform when compared to signalized and non-signalized intersections in lieu of real-world data. Zhou \textit{et al.} \cite{zhou_longfei} analyzed the performance of roundabouts compared to signalized intersections. They considered several metrics, including the mean number of vehicles passing through an intersection per unit time, mean speed of each vehicle, mean number of halts per vehicle, and mean time loss per vehicle. The authors defined a halt as speed passing below 1.39 m/s for longer than 1 second. Their results showed that roundabouts allow more vehicles to pass through per unit of time. However, roundabouts also cause more halts than signalized intersections especially as traffic scales.

Dabbagh \textit{et al.} \cite{al_dabbagh} used slightly different metrics to assess the performance of cities with cross-intersections, roundabouts, and both. They used a cumulative time that vehicles were completely stopped due to traffic. Furthermore, they assess performance based on the excess time a vehicle’s trip takes due to traffic. In our paper, we use a combination of these metrics to assess the performance of a topology, along with a new metric devised to fairly compare topologies, discussed in more detail in Section \ref{sec:metrics}.

\section{Zonal Road Topology}
\label{sec:topology}

%%Zonal Road Topology

\begin{comment}
The Zonal Road Topology is best illustrated by the simplified Figure \ref{fig:zonal-desc} of a two loop by two loop road system. 
This road topology does not contain any method of traditional traffic control like stop signs or traffic lights. 
The Zonal Road Topology necessitates cross-over merges, which are similar to zipper merges, but with two lanes as input to the merge, and two lanes as output. depicted in Figure \worry{Y}. \worry{Jason: Do you have any depictions of cross-over merges?}
It is meant for the exclusive use of fully autonomous vehicles, allowing for the seamless execution of cross-over merges,
without the hindrance of driver error. In the ideal case, traffic continuously flows and the transfer of loops is done by merging where the loops connect. Loops of road are unidirectional, meaning vehicles travelling on a given loop can only travel in the clockwise or counter-clockwise direction.
\end{comment}

\begin{figure}[b]
	\includegraphics[scale=0.3]{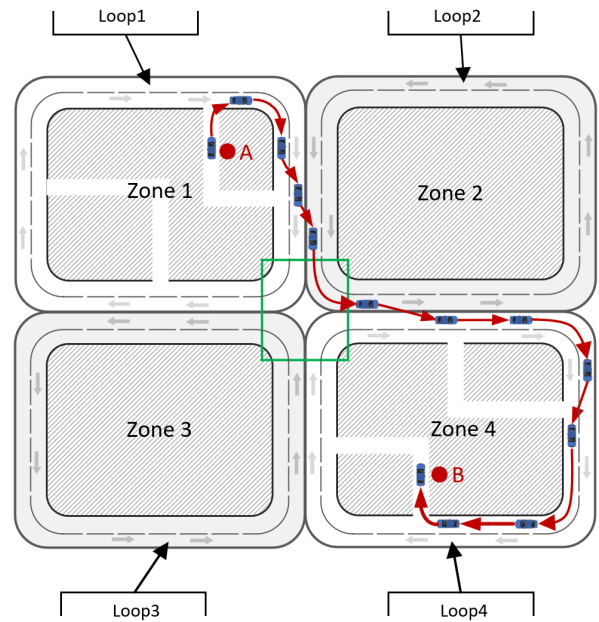}
	\centering
	\caption{Sample vehicle route in the Zonal Road Topology on a small grid.}
	\label{fig:zonal-desc}
\end{figure}

Fig. \ref{fig:zonal-desc} presents a simple visualization of the Zonal Road Topology, a network comprising a two-by-two loop configuration. Each self-contained zone is circumscribed by a loop: a unidirectional roadway free from classic traffic controls like stop signs and traffic lights. On each loop, traffic flows either clockwise or counter-clockwise which ensures a constant and predictable traffic flow. Vehicles navigate this system using designated crossover lanes that facilitate movement from one loop to another. Illustrated in the figure, the vehicle's path from point A to point B involves merging onto Loop 1 from a side street, transitioning through lanes to enter Loop 2 and thereafter Loop 4, before finally exiting onto a local street in Zone 4. This infrastructure is exclusively optimized for fully autonomous vehicles, ensuring uninterrupted flow and seamless loop transitions, thereby removing the risk of driver errors.

\begin{comment}
Traditional intersections contain 16 crossing conflict points, 8 diverging conflict points, and 8 merging conflict points. Roundabouts contain 4 merging, and 4 diverging conflict points. The Zonal Road Topology moves all conflict points to the straight sections, where there is only one type of conflict: merging, seen in Figure \ref{fig:conflicts} \worry{(We need to include in Figure 2 this same diagram but for Zonal)}. The simplification of decision making at conflict points will make the implementation of autonomous vehicles in our future cities much simpler.
%\worry{Do we need a citation for these sources of conflict?}
\end{comment}

\begin{figure}[t]
	\includegraphics[scale=0.14]{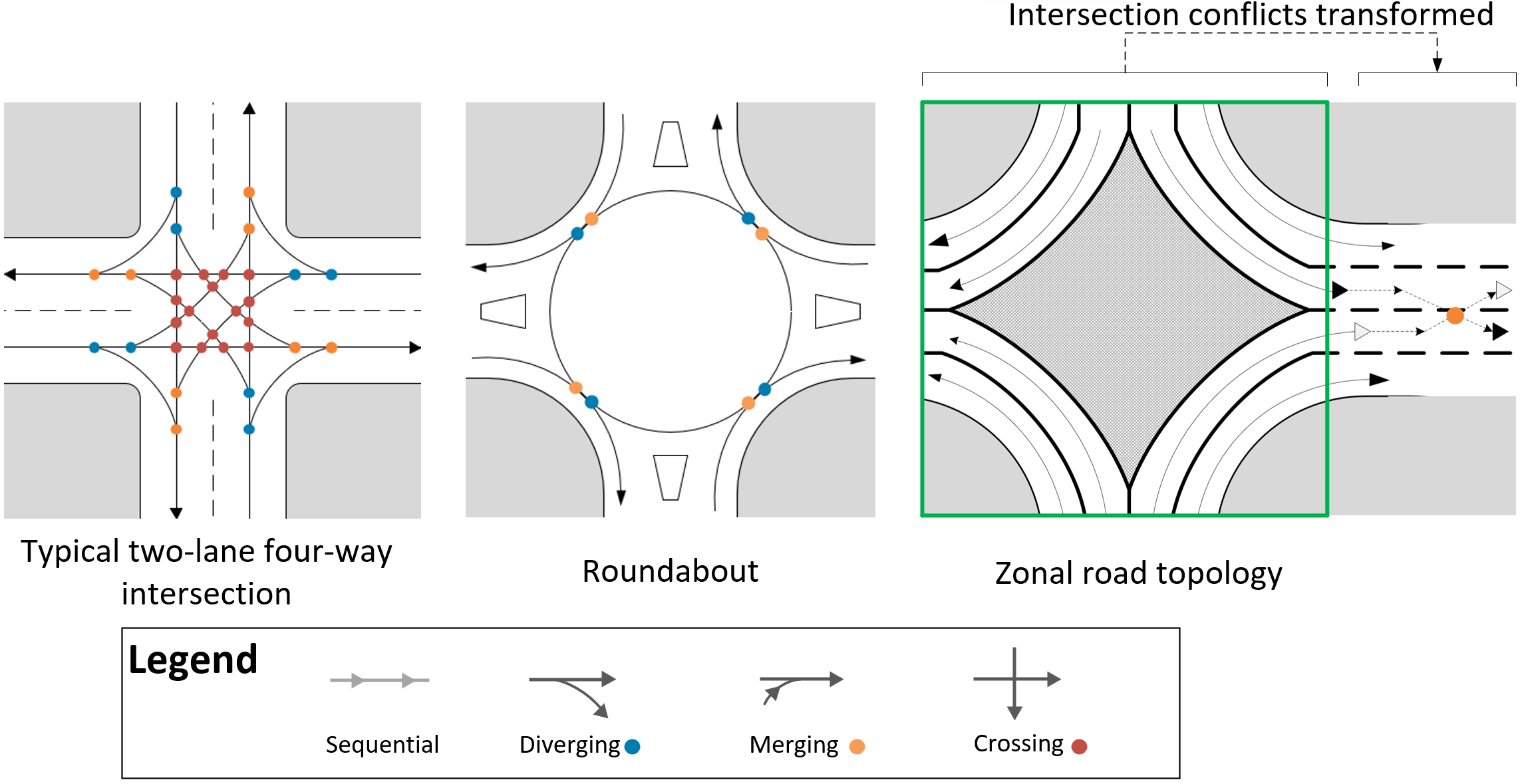}
	\centering
	\caption{Types of vehicular conflicts present in traditional cross-intersections, roundabouts, and the Zonal Road Topology.}
	\label{fig:conflicts}
\end{figure}

The green box area within the Zonal Road Topology of Fig.~\ref{fig:zonal-desc} departs from classic intersections by eliminating cross-traffic. This approach resolves classic vehicle conflicts such as crossing, merging, and diverging, which are common complexities in urban traffic design. Fig. \ref{fig:conflicts} demonstrates that a standard two-way traffic light intersection has 32 conflict points, whereas a roundabout has only 8\cite{SUDAS}. In contrast, the Zonal Road Topology intersection has none, because conflict points are reallocated to the road’s linear portions in the form of lane changes. The Zonal Road Topology thus simplifies the driving process by focusing mainly on merging. This simplification not only eases decision-making processes for autonomous driving systems, and hence simplifies data collection, but also effectively eliminates intersection accidents.

\begin{comment}
One of the major downsides to traffic-light based traffic control is the acceleration-deceleration cycles of traffic flow. Especially when the amount of traffic increases, these oscillatory cycles of stop-and-go traffic increase fuel consumption and emissions, and pose a safety hazard \cite{laval}. 
One of the prime supposed benefits of the Zonal Road Topology is that it allows for continuous traffic flow, eliminating the excessive time, energy, and emissions spent in these traffic oscillations. 
The elimination of traffic cycles also promises improvement to the amount of time spent driving, even though the distance traveled within the Zonal Road Topology would be on average further than the distance traveled on traditional roads.
\end{comment}

In the proposed Zonal Road Topology, pedestrians would not be present on the same plane as vehicles. Pedestrian traffic could be moved to a higher plane, such as between buildings, or on walkways built on top of the topology. Modelling pedestrian traffic along with vehicle traffic is not within the scope of this paper; instead, it will be deferred to future work.

\begin{figure}[b]
	\includegraphics[scale=0.2]{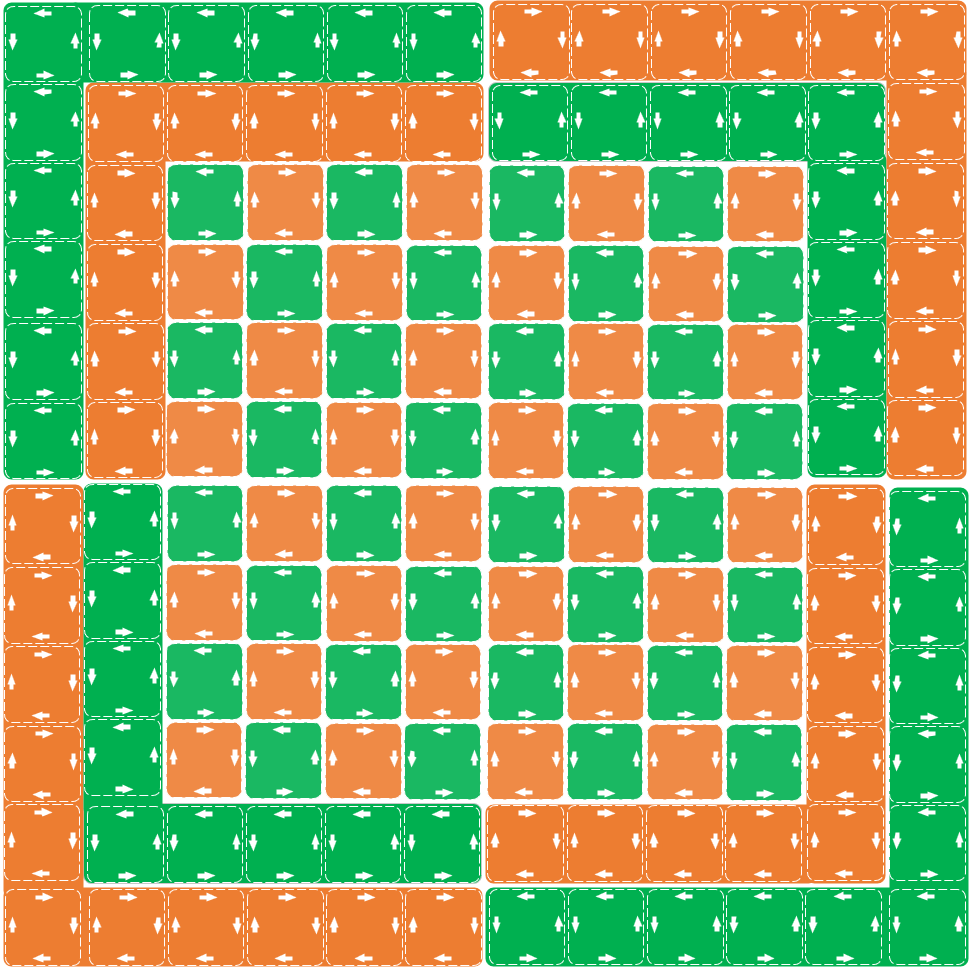}
	\centering
	\caption{Zonal Road Topology with a different configuration of clockwise (orange) and counter-clockwise (green) loops.}
	\label{fig:mosaickedloops}
\end{figure}

Fig. \ref{fig:mosaickedloops} illustrates a Zonal Road Topology design that more closely aligns with an actual urban configuration. At the core of the metropolis is the recurring loop-based pattern from Fig \ref{fig:zonal-desc}, aimed at reducing congestion in the downtown area. On the city's outskirts, elongated, narrow loops integrate basic loop structures, which are intended to expedite the movement of traffic moving longitudinally or laterally. This facilitates the rapid transitions between the northern and southern as well as the eastern and western corridors of the city.

\begin{comment}
Infrastructure such as buildings, homes, and charging stations are contained within zones, where the method of controlling traffic is traditional - with the use of stop signs, roundabouts, etc. Pedestrians would not be present on the same plane as vehicles in this topology. Pedestrian traffic could be moved to a higher plane, such as between buildings, or on walkways built on top of the topology. Modeling of pedestrian traffic along with vehicular traffic is not within the scope of this paper and will be future work. 

The Zonal Road Topology has the potential to be beneficial in many ways. First, it will be a continuously flowing smart city with the goal of having no traffic jams during busy periods like rush hour or in the case of events that result in unidirectional traffic. Second, it is safer, with the elimination of traditional cross-intersections with 32 conflict points. Third, it simplifies the decision process of autonomous driving systems by removing the “right at intersection”. It is a lower-resistance path toward the wide-scale implementation of L4 and L5 autonomous vehicles.
\end{comment}

One of the major downsides inherent in traffic-light-based road systems is the acceleration-deceleration cycles of traffic flow. Especially when the amount of traffic increases, these oscillatory cycles of stop-and-go traffic increase fuel consumption and emissions, and pose a safety hazard \cite{laval}. 
%These become particularly problematic as traffic volume increases, leading to a rise in fuel consumption, emissions, and potential safety risks. 
The Zonal Road Topology proposes a solution to this by facilitating a continuous traffic flow, where vehicles can maintain a constant speed throughout their journey. This consistency not only makes travel times and expected time of arrival (ETA) more predictable but also reduces the energy waste and emissions associated with stop-and-go traffic. Although the Zonal Road Topology might extend travel distances compared to traditional routes, the benefits of steady traffic flow could be substantial, particularly with the full integration of autonomous vehicles.

\begin{comment}
\item Zonal Topology allows for mosaicking of loop shapes, but that will be future work.
\item Zonal Topology has the potential to be beneficial in multiple ways. First, it will be a continuously flowing smart city, non traffic jam ever during busy periods like rush hour or unidirectional events. Second, it is a safer road network and eliminates traditional intersections for our future autonomous-vehicle cities. (Talk about intersection conflicts here) Third, simplifies the decision process of autonomous driving systems by removing the "right at intersection". Easier path, less resistance path toward L4 and L5. 
\item Natural question is where do pedestrians go? They go on a different plane.
\item Our analysis in simulation investigates a simple 10x10 grid, with no inner roads within zones.
%\end{itemize}
\end{comment}

\section{Experiment Setup}
\label{sec:methodology}

\subsection{Simulation Pipeline}
\begin{figure}[b]
	\includegraphics[scale=0.15]{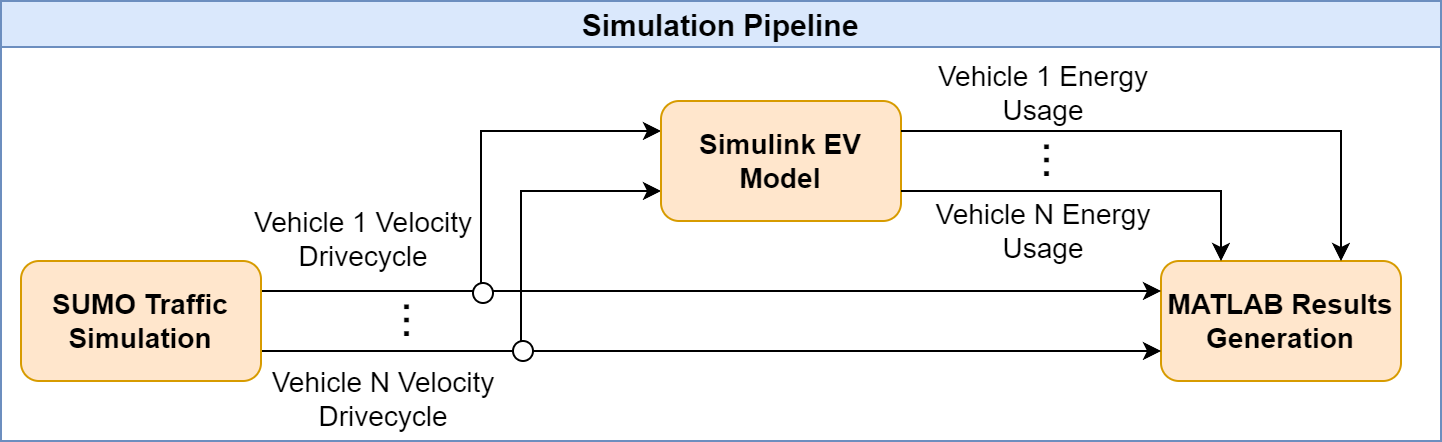}
	\centering
	\caption{Simulation pipeline used to run all the simulations in this paper.}
	\label{fig:pipeline}
\end{figure}

The experimental bench is shown in Fig.~\ref{fig:pipeline}, consisting of three main components.
The SUMO Traffic Simulation simulates vehicle traffic within each road topology to produce vehicle drive cycles. The Simulink EV Model models the powertrain of a generic EV to estimate energy consumption based on the drive cycle produced by SUMO. MATLAB Results Generation is a set of in-house MATLAB scripts that consume the raw data from SUMO and Simulink, and calculate and output the relevant metrics.

\subsection{Topology Generation}

The road topologies are modelled and simulated within the SUMO platform, an open-source software that allows for microscopic-level simulation of L5 vehicle traffic. SUMO provides the ability to model custom road networks, including both the traditional and Zonal Road topologies. Each topology in our experiments is a 10 x 10 grid of zones, with each zone being 250 m by 250 m in size and consisting of two lanes with a maximum road speed of 50 km/h. 

Within the traditional road topologies, we use static traffic lights which feature fixed-phase and timings, and adaptive traffic lights which adjust the timing of the lights based on the current state of traffic. This is accomplished using a delay-based method, which measures the accumulated time loss of vehicles entering the intersection. If any vehicle has an accumulated time loss greater than a threshold (set to one second for the simulation), a prolongation of the green light is requested unless the maximum green light duration of 45 seconds has been reached \cite{oertel}. To further increase the capacity of the road topologies, dedicated left turning lanes and lights are included within the traditional road topologies.

\begin{figure}[b]
	\includegraphics[scale=0.23]{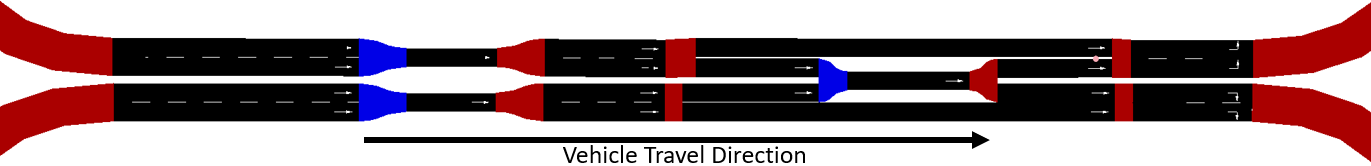}
	\centering
	\caption{Zipper merges within the Zonal Road Topology, highlighted in blue.}
	\label{fig:zipper_merges}
\end{figure}

For the Zonal Road Topology, the zone corners are modelled with curved roads of radius 30 m, with the maximum road speed on the curves maintained at 50 km/h. To simulate the lane changes required to switch between different zones, a series of zipper merges are implemented along the edges of a zone, as shown by the blue regions in Fig. \ref{fig:zipper_merges}. 
Although the zipper merges introduce additional slowdowns arising from choke points not present in the desired lane change, this provides finer control over lane changing behaviour within SUMO to avoid merging priority given to one lane and to specify the exact location of lane changes. Future work will consider more optimized lane merges.
%Although the zipper merges do not perfectly simulate the desired lane changes used within the Zonal Road Topology due to the vehicles slowing down when entering the zipper merge, this provides a consistent behaviour that enables vehicles to change zones.

\subsection{Traffic Generation}

Traffic is inserted into the simulation according to two distributions: a Poisson distribution and a spatial probabilistic distribution. 
The Poisson distribution is used to generate arrival times of new vehicles inserted into the road topology and is simulated across two hours in duration. 
Once a vehicle is set to be inserted into the simulation, the exact insertion road depends on the spatial probabilistic distribution. 
The spatial distribution is controlled through Origin Destination (OD) pairs, where the likelihood of the origin or destination road depends on weights assigned to the road based on the desired traffic flow pattern being simulated. 
Within this paper, a randomly distributed spatial distribution is used. 
Within the randomly distributed traffic, two main components of the traffic are included. The first component is a uniform weight across all roads. Then, to introduce noise into the system, 18 of the OD pairs, chosen at random, feature double the normal level of traffic to represent denser traffic areas within the road topology.

Once the road weights are established, varying levels of traffic are inserted into the road topologies using a unit of vehicles per hour per zone (vehicles/hour/zone) to determine the effects of varying traffic levels on the road topology's performance.
To further simplify the simulation, all inserted traffic consists of same sized sedans that behave identically in traffic.

\subsection{Electric Vehicle (EV) Model}
To analyze energy consumption, SUMO provides information about the energy consumption of a given vehicle. However, \cite{sagaama} showed that SUMO overestimates energy consumption by 3.03\% to 11.1\% when compared to real-world data provided by the Joint Research Centre of the European Commission, depending on the drive cycle. For this reason, we used a generic EV model from Simulink, which has been shown to compute energy consumption accurately when compared to a vehicle on a dynamometer \cite{adegbohun}. 

To simplify the analysis, an assumption exists that no energy losses are incurred by vehicles turning. However, previous work indicates that turning accounts for an average of 3.1\% of the energy consumption within electric buses \cite{camiel}.  

\subsection{Simulation}

For each traffic rate, 1000 random vehicles were analyzed within each simulation round and 24 simulation rounds were conducted for each road topology.
Due to the large number of compute hours required, parallelization of the simulations is conducted by executing the entire pipeline on compute clusters provided by the Digital Research Alliance of Canada.
The compute clusters consist of \textit{Intel Gold 6148 Skylake} CPUs at 2.4 GHz, and 18.4 CPU years were used to run simulations presented in this paper. 

\section{Metrics}
\label{sec:metrics}

% Basic metrics that do not really need an explanation
For a sample of vehicles logged in the simulation, the total drive time $t_{T}$ (s), distance travelled $d_{T}$ (m), total energy used $\mathit{E_{Used}}$ (MJ), average progress rate $\rho$ (\%), and speed deviation $\Delta v$ (\%) of each vehicle were logged. The number of vehicles present in the simulation, $N$, is measured as a time series.

\textit{Energy Used} ($\mathit{E_{Used}}$): Total energy used represents the amount of energy used throughout an entire drive cycle for a single vehicle in the simulation, and is defined as follows:
\begin{equation}
    \mathit{E_{Used}} = \mathit{E_{Output}} - \mathit{E_{Input}} + \mathit{E_{Lost}}
\end{equation}
where $\mathit{E_{Output}}$ is the energy used to drive the motors, $\mathit{E_{Input}}$ is energy gained from regenerative braking, and $\mathit{E_{Lost}}$ is energy lost to the environment, such as through friction. 

\begin{figure}[t]
	\includegraphics[scale=0.18]{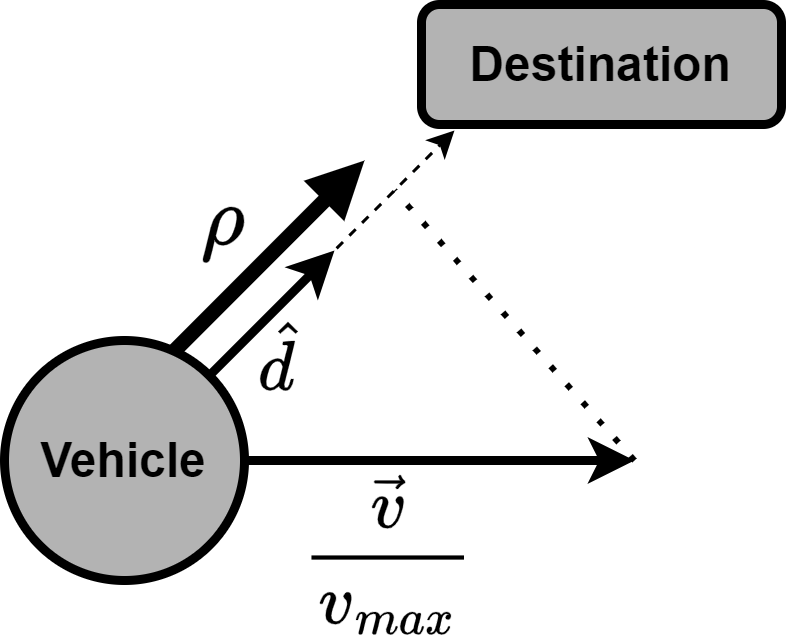}
	\centering
	\caption{A depiction of the Progress Rate metric. Progress Rate is maximized when $\vec{v}$ and $\hat{d}$ are in the same direction and $\lvert \vec{v}\rvert$ is equal to $v_{max}$.}
	\label{fig:progressRate_depiction}
\end{figure}

% Progress Rate
\textit{Progress Rate ($\rho$):} This metric evaluates the progress of a vehicle completing its trip over time. 
Its purpose is to illustrate how much of a trip is spent not making direct progress toward a vehicle's destination, due to low speeds or not travelling in the optimal direction toward the destination. Due to the routing that vehicles take within the Zonal Road Topology, vehicles generally do not make direct progress toward their destination, especially when the vector between the beginning and destination is horizontal or vertical. A lower progress rate can also occur in the traditional road topologies when, for example, vehicles are rerouted due to excessive traffic, or when vehicles are stopped either by traffic lights or excessive congestion.  Progress Rate is defined as follows:
\begin{equation}
    \rho \triangleq \frac{\vec{v}}{v_{max}} \cdot \hat{d}
\end{equation}
where $\hat{d}$ is the unit vector from the vehicle's current position to its target destination,  $\vec{v}$ is the velocity vector of the vehicle, and $v_{max}$ is the speed limit.
A visual representation of this is shown in Fig. \ref{fig:progressRate_depiction}.

This metric captures how much of the vehicle's instantaneous velocity contributes to reaching its destination. A vehicle with a velocity of $\vec{v}_{max}$, travelling directly towards the destination produces a $\rho$ of 1.0. When a vehicle's progress rate is 0, it indicates that the vehicle is stopped or travelling perpendicular to the destination. When a vehicle's progress rate is --1.0, the vehicle is travelling at $\vec{v}_{max}$ in the opposite direction to its destination.

%Speed Deviation
\textit{Speed Deviation ($\Delta v$):} To evaluate the magnitude by which a vehicle's speed deviates from the maximum road speed, we define $\Delta v$ as follows:

\begin{equation}
    \Delta v \triangleq \frac{v - v_{max}}{v_{max}}
\end{equation}
where $v$ is the vehicle speed and $v_{max}$ is the road's maximum speed. A $\Delta v$ of 0 represents a vehicle travelling at the road's maximum speed and $\Delta v < 0$ represents a vehicle travelling slower than the road's maximum speed. 
%$\overline{\Delta v}$ would be the mean speed deviation of a set of vehicles, and $\sigma \Delta v$ would be the standard deviation of the speed deviation for the same set. 

% Halts
\textit{Total Number of Halts:} A vehicle halt is defined as a vehicle travelling at a slow speed for longer than a threshold time. 
\begin{equation}
    Halt \iff  (\forall t | i \leq t \leq j \land j-i \geq \frac{2}{ts}: v(t) < v_H)
\end{equation}
where $ts$ is the time step, $v(t)$ is the speed of the vehicle as a function of time step, and $v_H$ is the halting speed which is set to 10 km/h in our simulations.
The halting time, \emph{i.\@e.\@}, the interval $[i, j]$, was set to 2 seconds. 

\begin{comment}
% Throughput
\textit{Throughput (T):} To help analyze the throughput of each topology, the $t_{C}$ is logged for each vehicle. 
Vehicles that do not complete within the maximum simulation time are not considered for this metric.
Throughput at time t, $T_{t}$, is generated by counting the number of vehicles that finishes at time t across all simulation rounds. 
\begin{equation}
    T_{t} = \mid\{\forall v | v \in Vehicles: v.t_{C} == t\}\mid
\end{equation}
where $T_{t}$ is the throughput at time $t$, $Vehicles$ is the set of all logged vehicles in the simulation that complete its trip, and $v.t_{C}$ is a vehicle's $t_{C}$.
\end{comment}

\textit{Number of Vehicles Present in the Simulation (N):} This metric is used to understand how many vehicles are present within the road topology at a given time step. An increasing value above the steady-state value indicates that the road topology is unable to allow vehicles to complete their trip in a timely manner due to excessive traffic. At steady-state, the number of vehicles completing their trip within a set time window will be near the number of vehicles being spawned. Thus, the net change in $N$ will be close to 0. When the topology has reached capacity, the number of vehicles being spawned will become greater than the number of vehicles finishing their trip, thus $N$ will increase. $N$ is measured as a time-series in contrast to the other metrics analyzed.

\section{Results}
\label{sec:results}

We will now analyze and discuss simulation results with respect to the metrics discussed above, sectioned into attributes that affect the quality of the average trip. In particular, 
Travel Efficiency, Safety, Energy Usage, and Topology Capacity will be discussed.

\subsection{Travel Efficiency}
% Drive Time
Simulations show that, on average, vehicles in the Zonal Road Topology travel 33\% longer distances than vehicles in the traditional road topologies.
This additional drive distance raises a concern as to whether the drive time is impacted.

\begin{figure*}[!t]
   \begin{subfigure}[T]{.33\textwidth}
        \centering
        \includegraphics[scale=0.43]{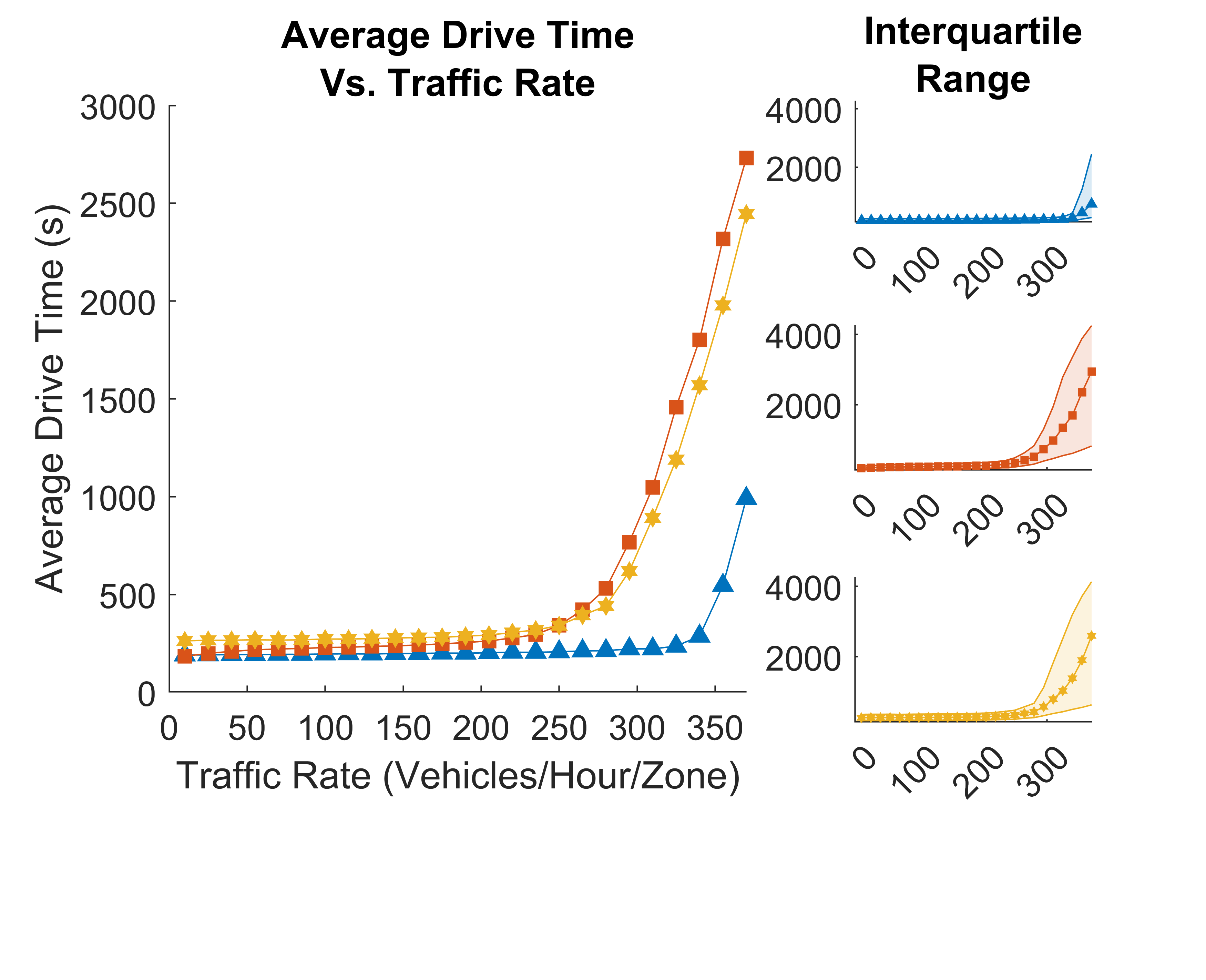}
        \caption{Average Drive Time.}
        \label{fig:driveTime}
    \end{subfigure}
    \begin{subfigure}[T]{.33\textwidth}
        \centering
        \includegraphics[scale=0.43]{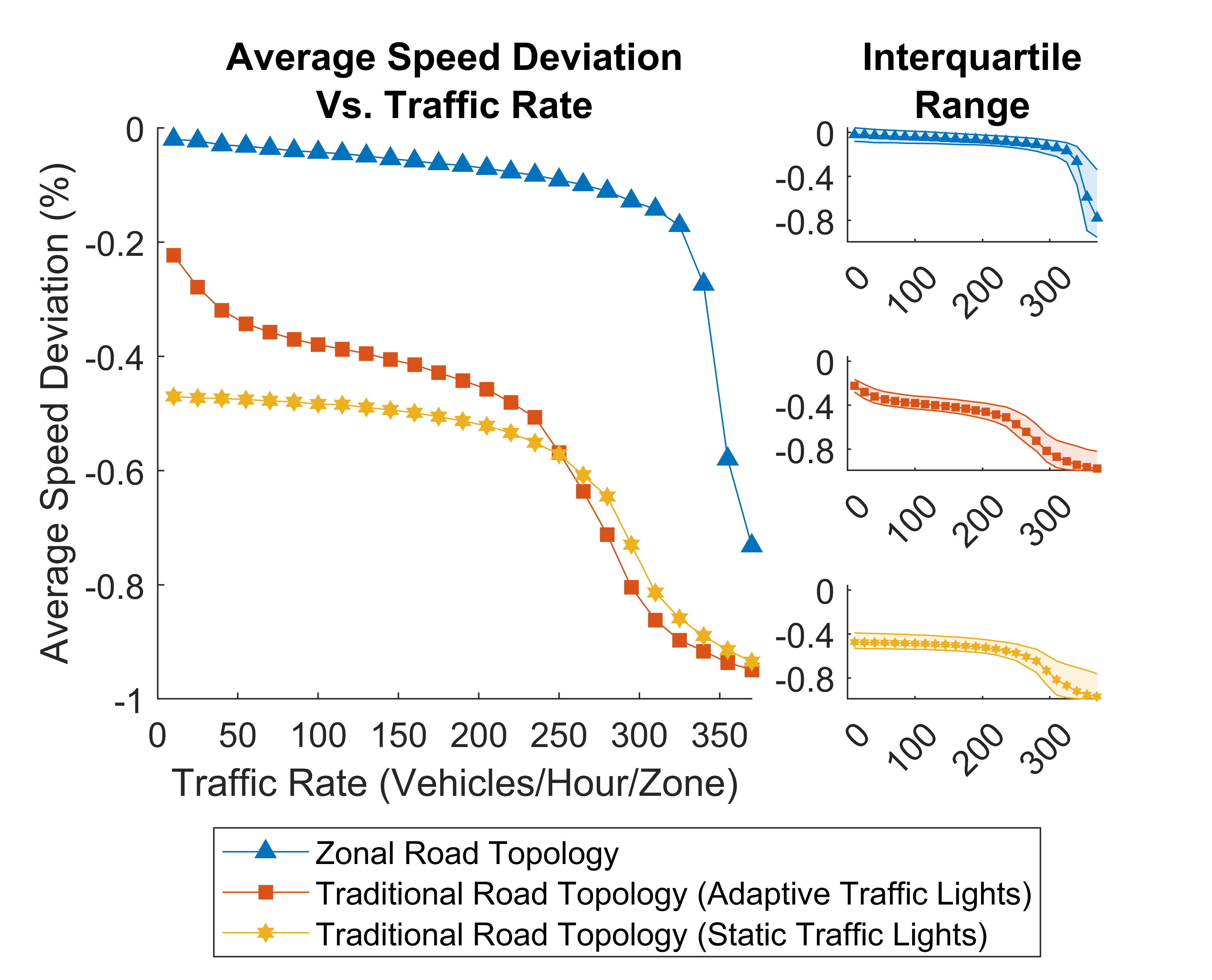}
        \caption{Average Speed Deviation.}
        \label{fig:speedDeviation}
        \end{subfigure}
   \begin{subfigure}[T]{.33\textwidth}
        \centering
        \includegraphics[scale=0.43]{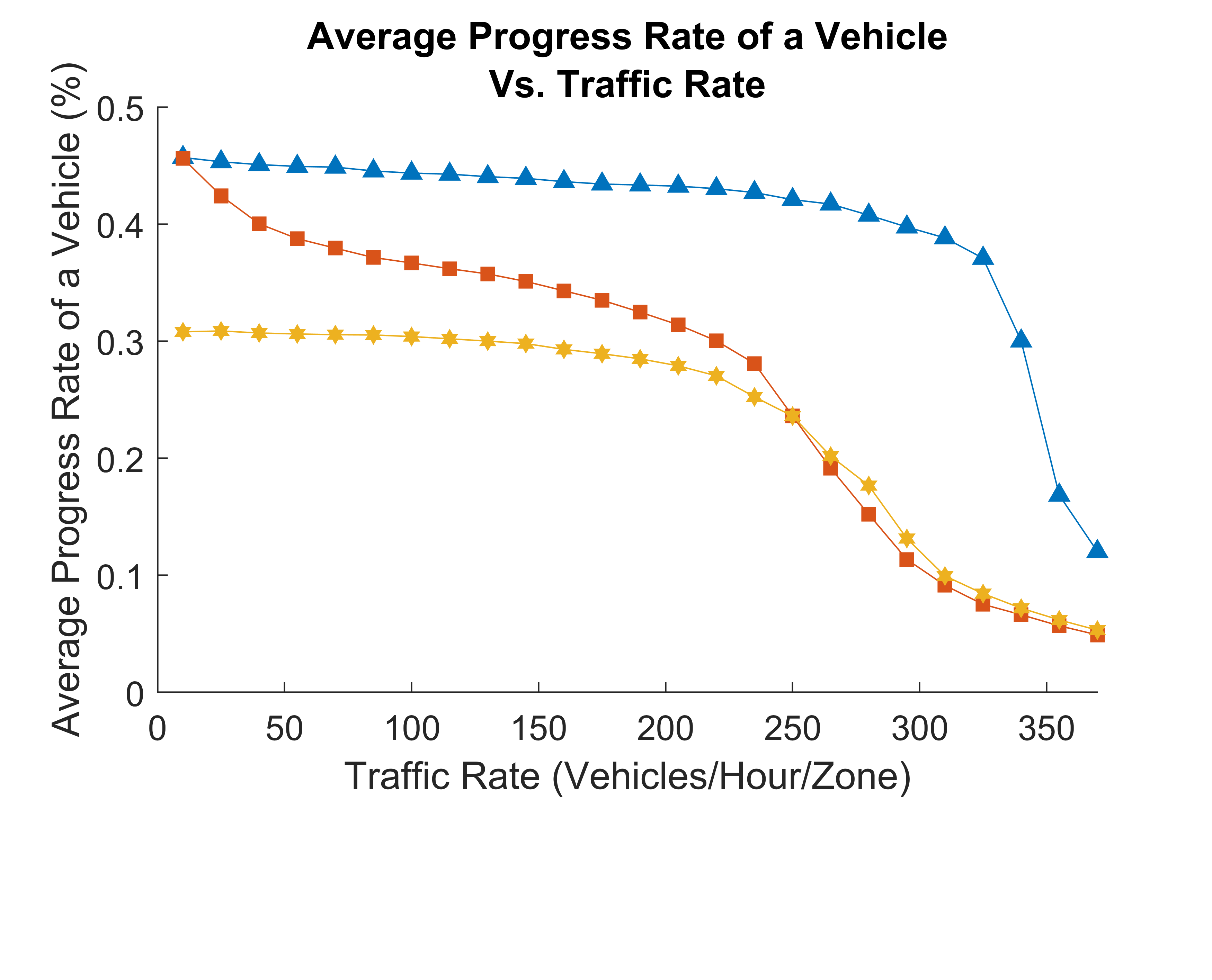}
    	\caption{Average Progress Rate.}
    	\label{fig:progressRate_results}
    \end{subfigure}
    \caption{Travel Efficiency of the Zonal Road Topology versus the traditional road topologies across varying levels of traffic. (a) The Zonal Road Topology has a lower, more consistent drive time. (b) The Zonal Road Topology allows vehicles to travel closer to the speed limit for longer. (c) A similar trend as in the average speed deviation is observed.}
    \label{fig:travelEfficiency}
\end{figure*}

Fig. \ref{fig:driveTime} shows that at low traffic rates, vehicles in the Zonal Road Topology have consistently lower drive times compared to the traditional road topologies, however small the benefit. As traffic starts to scale upward, the advantage that the Zonal Road Topology provides becomes sizeable.
It is clear from Fig. \ref{fig:driveTime} that all of the road topologies have a threshold traffic level after which the average drive time starts to rapidly increase. However, the Zonal Road Topology can support a greater traffic rate before experiencing this degradation. 
Prior to the inflection in drive time at 325 vehicles/hour/zone, the Zonal Road Topology has a minimal increase of 1 s of average drive time for every 10 vehicles/hour/zone increase. 
This indicates that the topology is minimally affected by different levels of traffic until a certain traffic level is reached, \emph{i.\@e.\@} the Zonal Road Topology exhibits a significantly higher level of resilience to traffic variability.
Furthermore, the Zonal Road Topology has a smaller Interquartile Range (IQR, a measure of statistical dispersion) as another indication that the topology is capable of providing consistent travel times to allow for better estimation of time to destination.
The increased consistency in drive time is due to a greater number of vehicles travelling close to, or at the speed limit for a greater duration of their trip, which is depicted in Fig. \ref{fig:speedDeviation}.

% Speed Deviation
The speed limit deviation of a vehicle's trip is influenced by two key factors. The first factor is traffic. The second factor relates to the attributes of the road topology, and how intersections are controlled.
As expected, the absolute value of speed deviation increases at higher traffic rates for all road topologies because there are more vehicles on the road. 
However, because the Zonal Road Topology does not force vehicles to wait at traffic lights for flow coordination, the average speed deviation will, on average, be closer to zero for Zonal Road Topologies than for traditional topologies, as evidenced by Fig. \ref{fig:speedDeviation}.

Another advantage of the Zonal Road Topology is that the average vehicle makes more progress on average in the journey toward its destination. 
Fig. \ref{fig:progressRate_results} shows that although the Zonal Road Topology requires vehicles to travel along the loops, which sometimes causes the vehicle to momentarily travel away from its destination, the Zonal Road Topology has a greater, more consistent progress rate across various traffic rates when compared to the traditional road topologies. Even though vehicles within the traditional road topologies never travel away from their destinations in our simulations, the vehicles make no progress when stationary at an intersection and experience greater speed deviations, resulting in a lower average progress rate.

% Progress Rate
\begin{comment}
\begin{figure}[t]
	\includegraphics[scale=0.6]{img/Results/TimeEfficiency/progressRate.png}
	\centering
	\caption{Average Progress Rate measured across various traffic injection rates.}
	\label{fig:progressRate_results}
\end{figure}
\end{comment}

%SAFETY
\subsection{Safety}

\begin{figure}[!b]
	\includegraphics[scale=0.6]{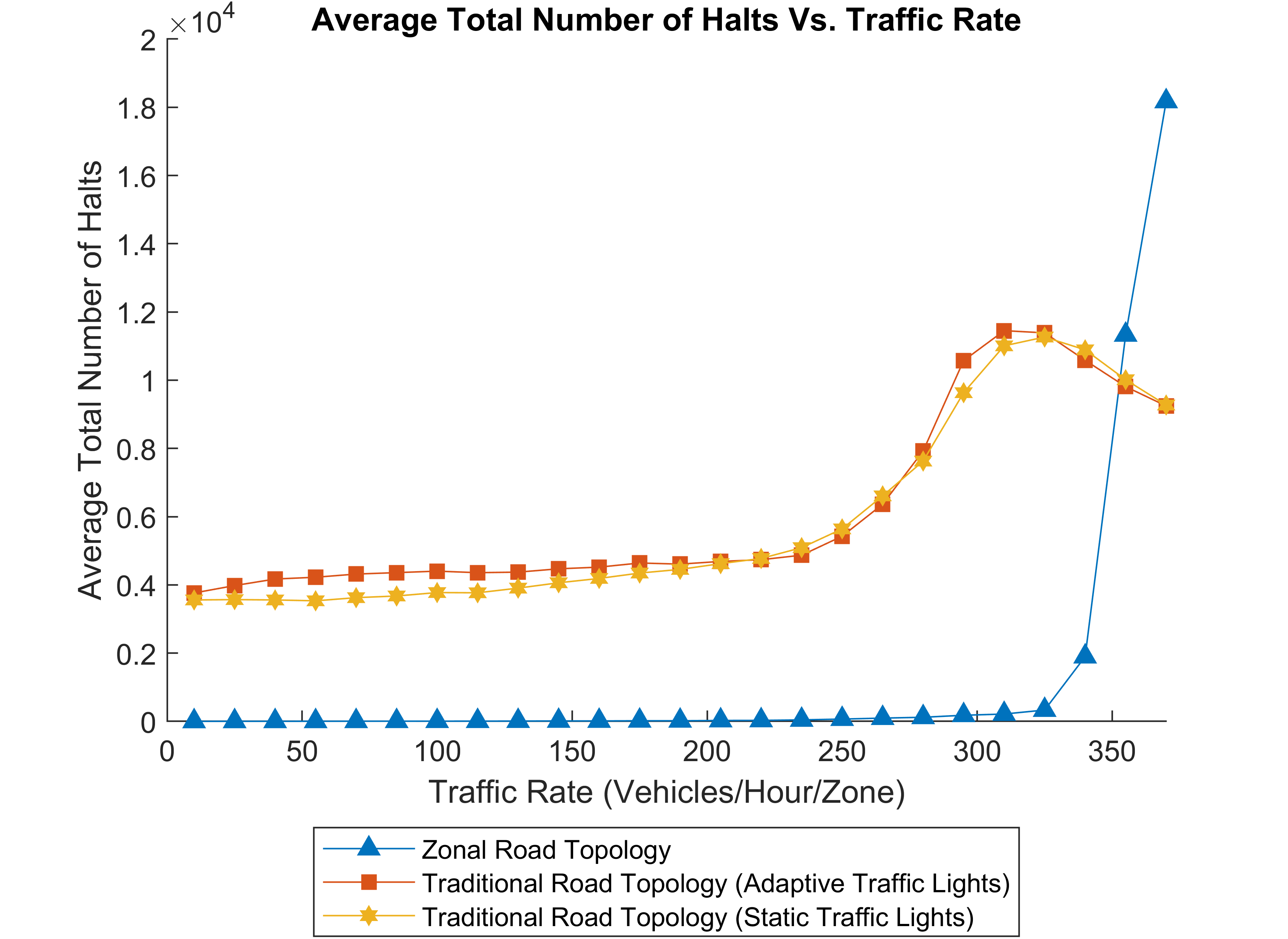}
	\centering
	\caption{Total number of halts experienced for all vehicles in the simulation, averaged across random seeds. }
	\label{fig:num_halts}
\end{figure}

 \begin{figure}[!b]
 	\includegraphics[scale=0.6]{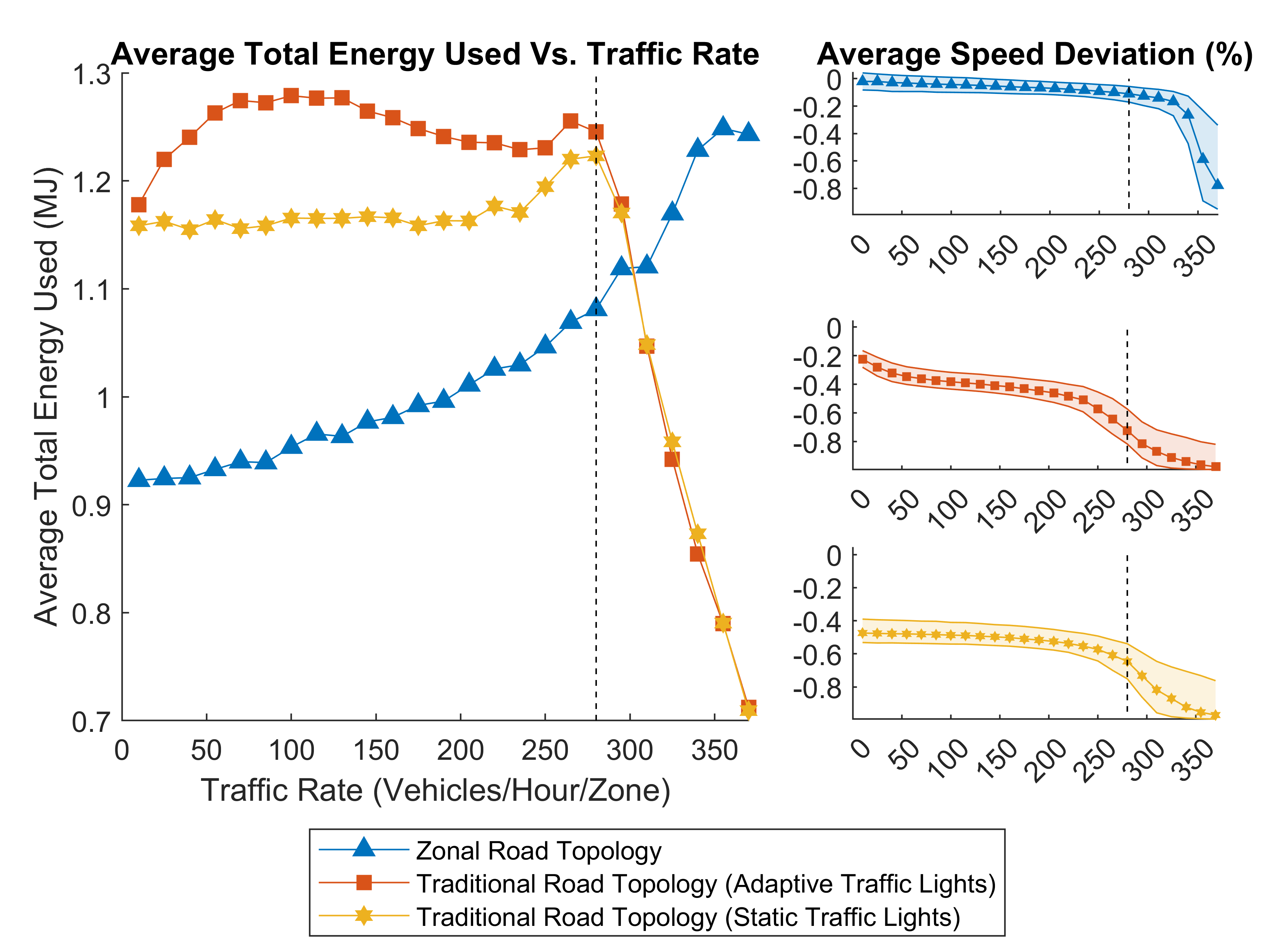}
 	\centering
 	\caption{(\textbf{Left}) Total energy usage, averaged across all vehicles. (\textbf{Right}) Average Speed Deviation. The point of decline is represented by the dashed vertical line.}
 	\label{fig:energy_usage}
 \end{figure}

\begin{figure*}[!bh]
   \begin{subfigure}[T]{.33\textwidth}
        \centering
        \includegraphics[scale=0.43]{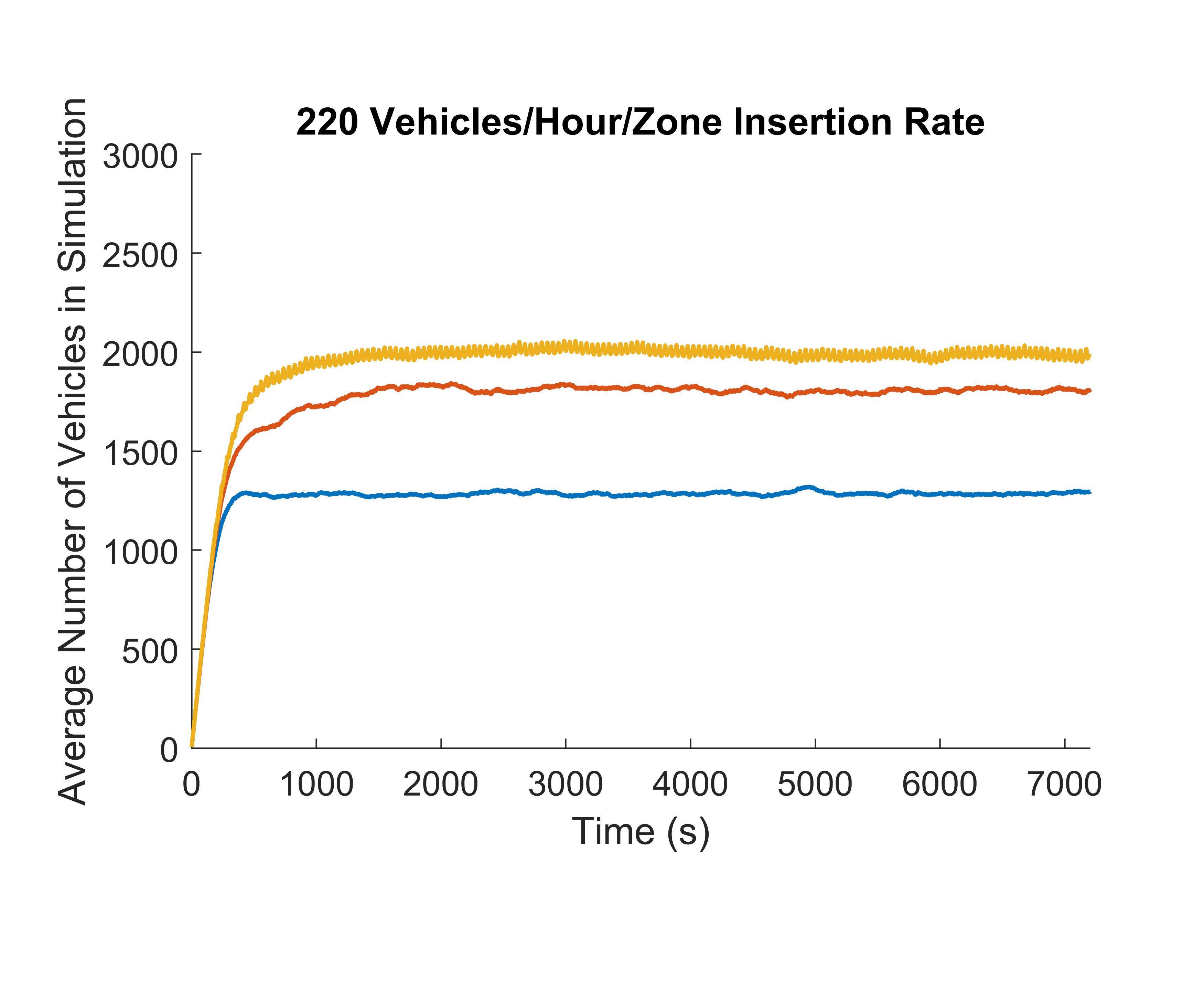}
        \caption{220 Vehicles/Hour/Zone.}
        \label{fig:numVeh_220}
    \end{subfigure}
    \begin{subfigure}[T]{.33\textwidth}
        \centering
        \includegraphics[scale=0.43]{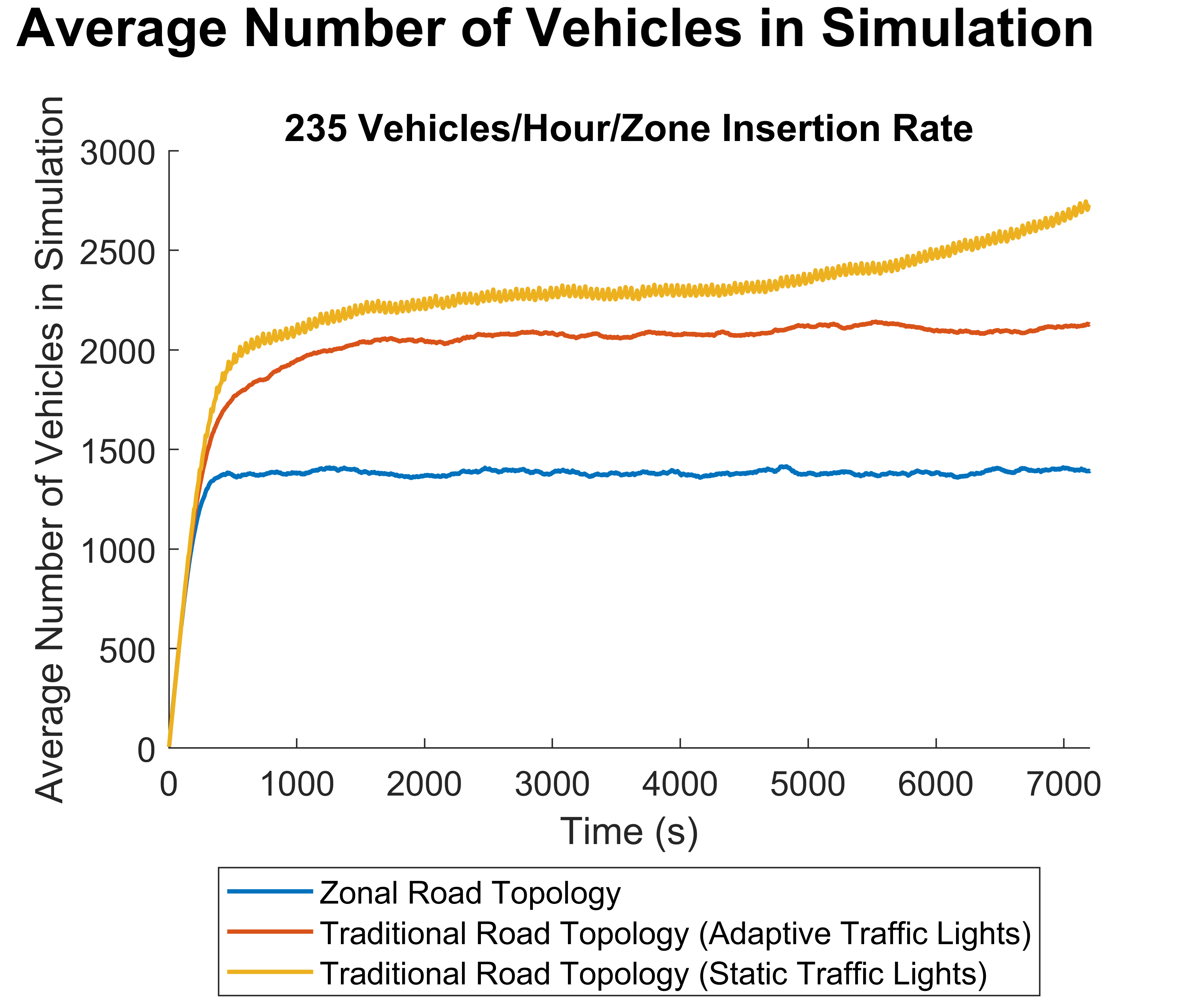}
        \caption{235 Vehicles/Hour/Zone.}
        \label{fig:numVeh_235}
        \end{subfigure}
   \begin{subfigure}[T]{.33\textwidth}
        \centering
        \includegraphics[scale=0.43]{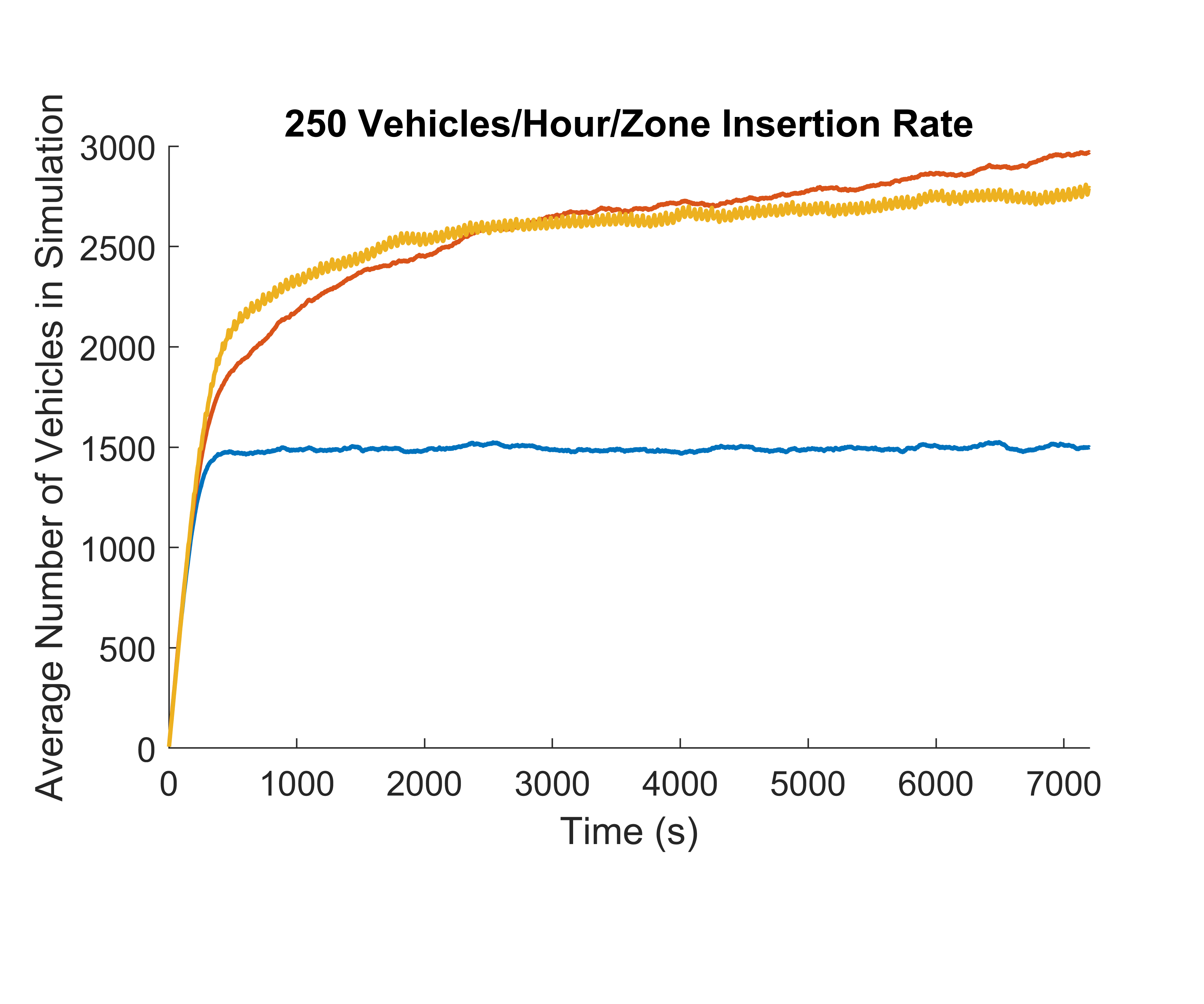}
    	\caption{250 Vehicles/Hour/Zone.}
    	\label{fig:numVeh_250}
    \end{subfigure}
    \caption{Number of vehicles in the simulation at each time step. (a) All topologies are able to handle traffic for the entirety of the simulation. (b) Traditional road topologies with static traffic lights inflect. (c) Traditional road topologies with adaptive traffic lights inflect.}
    \label{fig:trad_inflections}
\end{figure*}

The complex decision scenario present in traditional intersections not only makes the implementation of autonomous vehicles difficult, it also poses a safety hazard due to the multitude of stimuli a vehicle needs to be aware of in order to make a safe decision. Furthermore, the stop-and-go traffic oscillations inherent in traditional road topologies are undesirable to driver comfort and pose safety hazards. The appeal of the Zonal Road Topology is that vehicles halt significantly less frequently than in traditional road topologies for the majority of traffic levels, as seen in Fig. \ref{fig:num_halts}. The average number of halts present in the traditional road topologies inflects around 250 vehicles/hour/zone. However, the Zonal Road Topology continues to be able to support higher levels of traffic up to 340 vehicles/hour/zone when the frequency of halts increases drastically, surpassing the number of halts in the traditional road topologies. This is due to traffic in the Zonal Road Topology exhibiting stop-and-go behaviour when traffic is very dense. Meanwhile, in the traditional road topologies, the total number of halts begins to decrease past 310 vehicles/hour/zone due to traffic being so prohibitive as to cause near deadlock. The total number of halts has decreased simply because the duration of each halt is longer, which is evident in the large drive times seen in Fig.\@ \ref{fig:driveTime}.

\subsection{Energy Usage}

As discussed previously, vehicles within the Zonal Road Topology travel on average 33\% longer distances than vehicles within traditional road topologies. 
However, Fig. \ref{fig:energy_usage} shows that the Zonal Road Topology offers decreased energy usage for EVs despite the longer distances travelled, due to more consistent vehicle speeds, and hence, less acceleration and deceleration. 
 
A point of intersection between the energy curves of traditional and Zonal Road Topologies occurs at a traffic rate of around 300 vehicles/hour/zone, suggesting that the Zonal Road Topology does not handle higher traffic densities as well. 
However, further analysis reveals that this is not the case. Analyzing the energy usage in the traditional road topologies, a consistent decline in energy usage begins occurring at a rate of 280 vehicles/hour/zone. 
The cause of this decline can be explained in the average speed deviation plots within Fig. \ref{fig:energy_usage}. 
At the point of decline of 280 vehicles/hour/zone, the Zonal Road Topology has an average speed deviation of --11\%, in contrast to --71\% and --65\% for the adaptive and static traffic lights respectively. 
Therefore, vehicles within the traditional road topologies are traveling significantly slower due to heavy traffic congestion and cover less distance overall if the vehicle is unable to complete its trip within the maximum simulation time of two hours. 
This traffic congestion would superficially decrease the average energy usage of the congested vehicles, leading to the point of intersection seen within Fig. \ref{fig:energy_usage}.
The drop in energy usage being superficial is further evidenced by the drastic decrease in energy usage of the traditional road topologies after the point of decline, where the speed deviation quickly approaches --100\% which would indicate that the system has experienced a full deadlock. 
A similar trend begins within the Zonal Road Topology at the significantly higher traffic rate of 370 vehicles/hour/zone.

\subsection{Topology Capacity}

Due to the improvement in traffic flow that the Zonal Road Topology provides, it is expected that it can maintain a higher capacity of vehicles before the road system begins to suffer from excessive traffic. In this section, we identify the capacity of each road topology -- the traffic insertion rate at which the system experiences instability, that is, the road system is unable to support the amount of vehicles in the simulation, and excessive traffic begins to build up. We measure the number of vehicles present in the simulation ($N$) as a time-series. If $N$ begins to increase after reaching steady-state, drivers are spending more time in traffic, and the road system's throughput is declining. Fig. \ref{fig:numVeh_220} shows that $N$ reaches a steady state early into the simulation at 220 vehicles/hour/zone for all road topologies, indicating that there is no issue with traffic buildup yet.

Slightly increasing the traffic injection rate to 235 vehicles/hour/zone in Fig. \ref{fig:numVeh_235}, we see that the traditional road topology with static traffic lights becomes unstable, as evidenced by the increase in $N$ starting at around 4500 seconds. This is caused by a traffic jam, which prevents vehicles from exiting the simulation within the fixed simulation time. 

Soon after, at 250 vehicles/hour/zone, as seen in Fig. \ref{fig:numVeh_250}, the traditional road topology with adaptive traffic lights becomes unstable. Thus, we can conclude that the maximum capacity for the traditional road topology is approximately 235 vehicles/hour/zone if using static traffic lights, and 250 vehicles/hour/zone if using adaptive traffic lights.

\begin{figure}[t]
	\includegraphics[scale=0.55]{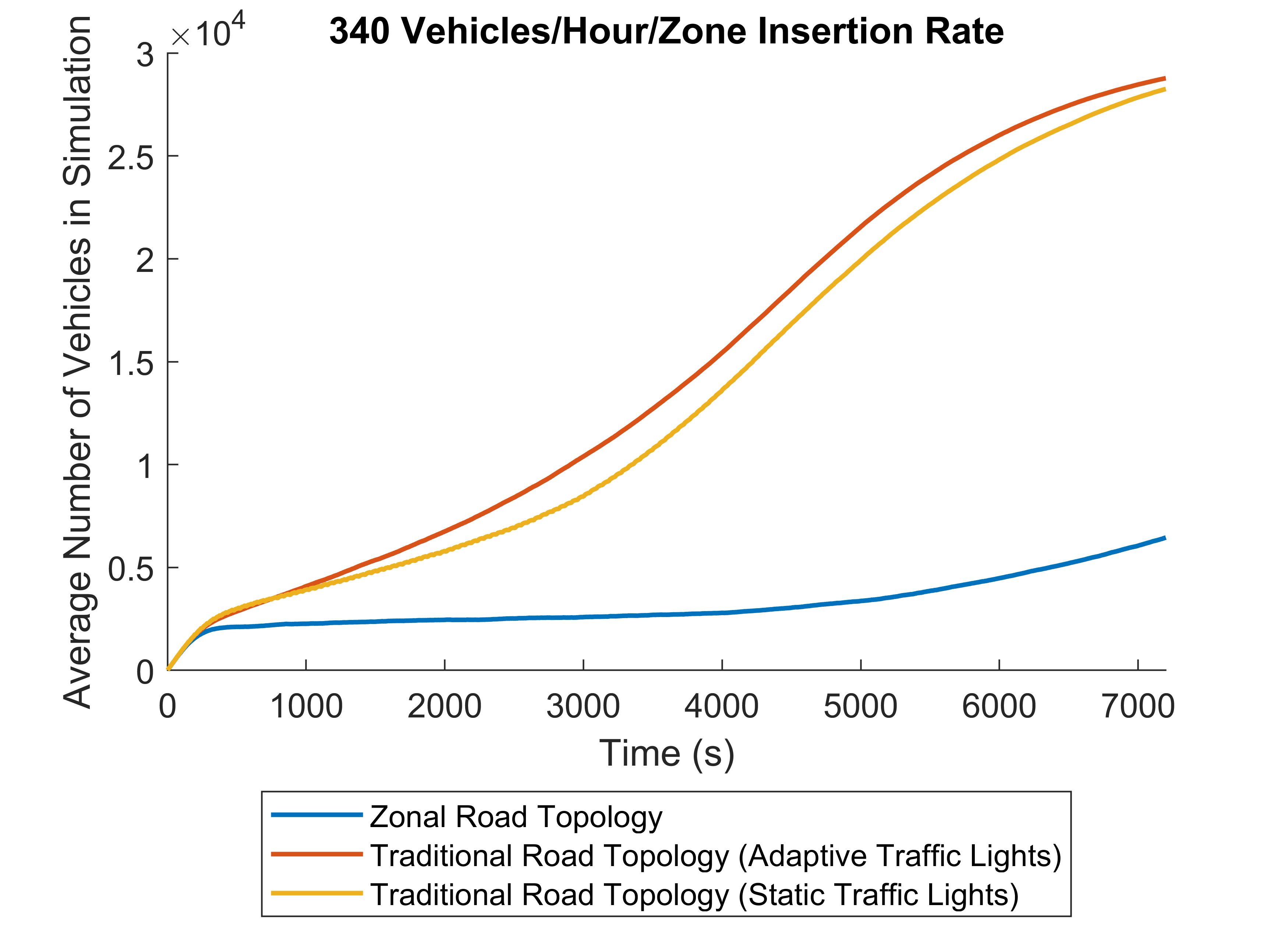}
	\centering
	\caption{Number of vehicles in the simulation at each time step. The Zonal Road Topology becomes unable to handle this level of traffic.}
	\label{fig:numVeh_340}
\end{figure}

The Zonal Road Topology continues to remain stable until 340 vehicles/hour/zone as seen in Fig. \ref{fig:numVeh_340}. Therefore, the maximum capacity for the Zonal Road Topology is around 325 vehicles/hour/zone. 

\begin{comment}
We define the drift to be the surplus of vehicle arrivals into the road topology as opposed to the vehicle departures. We expect that in steady state, the $arrivals - departures$ would oscillate around 0. When a simulation incurs too much traffic and a traffic jam occurs, this value will increase, and the road system will become unable to retire vehicles in time for new vehicles to spawn due to the traffic jam. Figure \ref{fig:lambda_280} shows the traditional road topologies becoming unstable, while the Zonal Road Topology is able to maintain a steady state at the same traffic density of 280 veh/hour/zone. At approximately 4000 seconds, the traditional road topologies become asymptotically unstable as the system reaches capacity.
\end{comment}

\section{Conclusions \& Future Work}
\label{sec:conclusion}
% conclusion
\begin{comment}
	- SUMO simulations with EV models
	- results show benefits: time+distance+speed, safety, energy, capacity
	- step toward goal of continuously flowing smart city and facilitates L4/L5 autonomy
	- to implement on a wide scale, would require collaboration between engineers, politicians, architetects, and lawyers
\end{comment}

In this study, we simulated the novel Zonal Road Topology using SUMO to compare its performance against the traditional road topologies that use traffic lights as intersection control. 
Our results show that the Zonal Road Topology offers decreased travel times, energy usage, and a greater average speed. 
Furthermore, the Zonal Road Topology has a greater topology capacity, allowing for higher levels of traffic before excessive traffic starts to build up. 
The implementation of this new road topology would bring us one step closer toward a continuously flowing smart city, and can also facilitate the wide-scale adoption of L4/L5 autonomy. 

\begin{comment}
This paper demonstrated the benefits of the Zonal Road Topology as it \suggestion{is safer, and more time, speed, energy and network-efficient} compared to the Traditional Road Topologies.
Ultimately, it shows that there is potential to improve the current traffic infrastructure; however, more work is needed as the Zonal Road Topology can not be simply deployed onto current city layouts.
\worry{Why can't it? We shouldn't introduce new ideas in the conclusion. And I don't think we discussed implementation anywhere yet. And is this about physically implementing the roads, or the two points discussed in the next sentence?} 
A higher level of vehicle autonomy is required and traffic laws need to be modified, which can take a long time to implement. 
\worry{Yes, we need more vehicle autonomy but isn't the zonal topology supposed to help the development of that? If we already have L5, then one of the benefits of zonal is lost.}
Effective city planning on a Zonal Road blueprint \suggestion{Topology} will bring society a step closer to an achievable smart city.
\worry{What is a Zonal Road blueprint? We didn't use this term before.} 
The future of smart cities must have a safe, fast, and balanced transportation system to move citizens in an efficient algorithm. 
However, it requires a team of engineers of all fields, politicians, architects, and lawyers to come together and refactor the existing infrastructure.
\end{comment}

% Future work
Future work will include an investigation into the performance of the Zonal Road Topology with various topology configurations to investigate its potential under varying traffic flow patterns and city shapes. 
%Some configurations include injecting the traditional road topologies inside the zones of the Zonal Road Topology, as this was the initial intent of the design. 
This includes simulating the internals of each zone, which will contain traditional methods of intersection control like those shown in Fig. \ref{fig:zonal-desc}.
Further experimentation will analyze various sizes of networks and zones, as well as mosaicking different zone shapes together, like those in Fig.~\ref{fig:mosaickedloops}.
%The goal would be to gather information from our experiments to develop a guideline for designing better smart cities. 
Our overarching goal is to understand how to efficiently implement the Zonal Road Topology in cities around the world in a way that will reduce congestion and improve energy efficiency.
To reach this goal, our team will need to perform higher fidelity simulations by allowing simulated vehicles to use Vehicle-to-Everything (V2X) information, which allow for platooning and improved cooperation between vehicles.
In parallel, the team is using 1/24 scale autonomous vehicles in a Vehicle-in-the-Loop setup to emulate traffic on the Zonal Road Topology to validate its benefits in hardware. Finally, we plan to further investigate the safety benefits and concerns of the Zonal Road Topology in comparison with traditional road topologies.

% \section{Acknowledgment}
% The research project is funded by the Natural Sciences and Engineering Research Council of Canada (NSERC) and the Federal Economic Development Agency for Southern Ontario (FedDev Ontario).

\bibliographystyle{ieeetr}
\bibliography{references}

\begin{thebibliography}{10}

\bibitem{martinezdiaz}
M.~Martínez-Díaz and F.~Soriguera, ``Autonomous vehicles: theoretical and practical challenges,'' {\em Transportation Research Procedia}, vol.~33, pp.~275--282, 2018.
\newblock XIII Conference on Transport Engineering, CIT2018.

\bibitem{AAA}
``Automatic emergency braking with pedestrian detection,'' tech. rep., American Automobile Association, Inc., 2019.

\bibitem{choi}
E.-H. Choi, {\em Crash Factors in Intersection-Related Crashes: An On-Scene Perspective}.
\newblock U.S. Department of Transportation, National Highway Traffic Safety Administration, 2010.

\bibitem{loos}
S.~M. Loos and A.~Platzer, ``Safe intersections: At the crossing of hybrid systems and verification,'' in {\em 2011 14th International IEEE Conference on Intelligent Transportation Systems (ITSC)}, pp.~1181--1186, 2011.

\bibitem{shi}
M.-k. Shi, H.~Jiang, and S.-h. Li, ``An intelligent traffic-flow-based real-time vehicles scheduling algorithm at intersection,'' in {\em 2016 14th International Conference on Control, Automation, Robotics and Vision (ICARCV)}, pp.~1--5, 2016.

\bibitem{azimi}
R.~Azimi, G.~Bhatia, R.~R. Rajkumar, and P.~Mudalige, ``Stip: Spatio-temporal intersection protocols for autonomous vehicles,'' in {\em 2014 ACM/IEEE International Conference on Cyber-Physical Systems (ICCPS)}, pp.~1--12, 2014.

\bibitem{elvik}
R.~Elvik, ``Road safety effects of roundabouts: A meta-analysis,'' {\em Accident Analysis \& Prevention}, vol.~99, pp.~364--371, 2017.

\bibitem{demir}
H.~Göçmen~Demir and Y.~K. Demir, ``A comparison of traffic flow performance of roundabouts and signalized intersections: A case study in nigde,'' {\em The Open Transportation Journal}, vol.~14, pp.~120--132, 07 2020.

\bibitem{zhou_longfei}
L.~Zhou, L.~Zhang, and C.~S. Liu, ``Comparing roundabouts and signalized intersections through multiple- model simulation,'' {\em IEEE Transactions on Intelligent Transportation Systems}, vol.~23, no.~7, pp.~7931--7940, 2022.

\bibitem{ARIUS}
K.~Salamati, N.~Rouphail, C.~Frey, and B.~Schroeder, {\em Accelerating Roundabouts in the U.S.: Volume III of VII - Assessment of the Environmental Characteristics of Roundabouts}.
\newblock Federal Highway Administration, 2015.

\bibitem{Jen2024}
D.~Jen, ``Dedicated self-driving truck lane to open in japan,'' {\em Civil Engineering Magazine}, January 2024.

\bibitem{he2022}
S.~He, F.~Ding, C.~Lu, and Y.~Qi, ``Impact of connected and autonomous vehicle dedicated lane on the freeway traffic efficiency,'' {\em European Transport Research Review}, vol.~14, 12 2022.

\bibitem{chen2022}
Y.~Chen, H.~Zhang, D.~Wang, and J.~Wang, ``Overall influence of dedicated lanes for connected and autonomous vehicles on freeway heterogeneous traffic flow,'' {\em Journal of Advanced Transportation}, vol.~2022, pp.~1--15, 08 2022.

\bibitem{patent}
X.~Hu, ``One-way loop mosaicking for higher transportation capacity and safety,'' U.S. Patent 0404123, Dec. 30, 2021.

\bibitem{duarte}
F.~Duarte, ``Self-driving cars: A city perspective,'' {\em Science Robotics}, vol.~4, 03 2019.

\bibitem{sumo}
P.~A. Lopez, M.~Behrisch, L.~Bieker-Walz, J.~Erdmann, Y.-P. Flötteröd, R.~Hilbrich, L.~Lücken, J.~Rummel, P.~Wagner, and E.~Wiessner, ``Microscopic traffic simulation using sumo,'' in {\em 2018 21st International Conference on Intelligent Transportation Systems (ITSC)}, pp.~2575--2582, 2018.

\bibitem{al_dabbagh}
M.~S.~M. Al-Dabbagh, A.~Al-Sherbaz, and S.~Turner, ``The impact of road intersection topology on traffic congestion in urban cities,'' in {\em Intelligent Systems and Applications} (K.~Arai, S.~Kapoor, and R.~Bhatia, eds.), (Cham), pp.~1196--1207, Springer International Publishing, 2019.

\bibitem{SUDAS}
SUDAS, {\em SUDAS Design Manual - 2023 Edition}, ch.~5.
\newblock Statewide Urban Design and Specifications Program, 2023.

\bibitem{laval}
L.~J.A. and L.~J., ``A mechanism to describe the formation and propagation of stop-and-go waves in congested freeway traffic,'' {\em Philosophical Transactions of The Royal Society}, vol.~368, 2010.

\bibitem{oertel}
R.~Oertel and P.~Wagner, ``Delay-time actuated traffic signal control for an isolated intersection,'' in {\em Proceedings 90st Annual Meeting Transportation Research Board (TRB)}, 01 2011.

\bibitem{sagaama}
I.~Sagaama, A.~Kchiche, W.~Trojet, and F.~Kamoun, ``Evaluation of the energy consumption model performance for electric vehicles in sumo,'' in {\em 2019 IEEE/ACM 23rd International Symposium on Distributed Simulation and Real Time Applications (DS-RT)}, pp.~1--8, 2019.

\bibitem{adegbohun}
F.~Adegbohun, A.~von Jouanne, B.~Phillips, E.~Agamloh, and A.~Yokochi, ``High performance electric vehicle powertrain modeling, simulation and validation,'' {\em Energies}, vol.~14, no.~5, 2021.

\bibitem{camiel}
C.~J. Beckers, I.~J. Besselink, and H.~Nijmeijer, ``Assessing the impact of cornering losses on the energy consumption of electric city buses,'' {\em Transportation Research Part D: Transport and Environment}, vol.~86, p.~102360, 2020.

\end{thebibliography}

\end{document}